\PassOptionsToPackage{table,xcdraw}{xcolor}

\documentclass[biblatex,english]{lni}
\AtEveryBibitem{
    \ifboolexpr{
        not test {\ifentrytype{online}} and
        not test {\ifentrytype{software}}
      }
        {\clearfield{url}}
    {}
}
\usepackage{booktabs}
\usepackage{graphicx}

\usepackage[]{blindtext}

\begin{document}

\newbibmacro{string+url}[1]{%
\iffieldundef{url}{
  \iffieldundef{doi}{#1}{\href{https://dx.doi.org/\thefield{doi}}{#1}}%
  }{\href{\thefield{url}}{#1}}}
\DeclareFieldFormat*{title}{\usebibmacro{string+url}{#1}}

\title[KnoWoGen]{Using Large Language Models to Generate Authentic Multi-agent Knowledge Work Datasets}

\author[1]{Desiree Heim}{desiree.heim@dfki.de}{0000-0003-4486-3046}
\author[2]{Christian Jilek}{christian.jilek@dfki.de}{0000-0002-5926-1673}
\author[3]{Adrian Ulges}{adrian.ulges@hs-rm.de}{}
\author[1]{Andreas Dengel}{andreas.dengel@dfki.de}{0000-0002-6100-8255}%
\affil[1]{Smart Data and Knowledge Services Department, German Research Center for Artificial Intelligence (DFKI) and Department of Computer Science, University of Kaiserslautern-Landau (RPTU)}
\affil[2]{Smart Data and Knowledge Services Department, German Research Center for Artificial Intelligence (DFKI)}
\affil[3]{Department DCSM, RheinMain University of Applied Sciences}
\maketitle

\begin{abstract}
Current publicly available knowledge work data collections lack diversity, extensive annotations, and contextual information about the users and their documents. These issues hinder objective and comparable data-driven evaluations and optimizations of knowledge work assistance systems. Due to the considerable resources needed to collect such data in real-life settings and the necessity of data censorship, collecting such a dataset appears nearly impossible. For this reason, we propose a configurable, multi-agent knowledge work dataset generator. This system simulates collaborative knowledge work among agents producing Large Language Model-generated documents and accompanying data traces. Additionally, the generator captures all background information, given in its configuration or created during the simulation process, in a knowledge graph. Finally, the resulting dataset can be utilized and shared without privacy or confidentiality concerns. 

This paper introduces our approach's design and vision and focuses on generating authentic knowledge work documents using Large Language Models. Our study involving human raters who assessed 53\% of the generated and 74\% of the real documents as realistic demonstrates the potential of our approach. Furthermore, we analyze the authenticity criteria mentioned in the participants' comments and elaborate on potential improvements for identified common issues. 
\end{abstract}


\section{Introduction}
Conducting data-driven evaluations of knowledge work support tools allows for deeper, objective insights into the tools' weaknesses, strengths, and their reasons compared to user studies. Moreover, it enables meaningful comparisons that are reproducible from the input data perspective. 

However, currently, there are no comprehensive knowledge work datasets available that fully satisfy desirable properties. Gon{\c{c}}alves~\cite{gonccalves2011pseudo} addressed this problem more than ten years ago and identified the incompleteness, lack of background information about the users and their documents, limited generalization, and the lack of ground truth information to base evaluations on as key issues of existing datasets. To address these issues, he envisioned a large-scale data collection from a representative set of users who must annotate their data extensively. However, he also mentioned the difficulty of realizing this vision.

A recent publication of Bakhshizadeh et al.~\cite{BakhshizadehJilekSchroeder2024} shows that these problems persist. They introduced a data collection named RLKWiC addressing the issues of missing background information by having eight participants track and annotate their daily work for two months and categorizing it into self-defined and annotated contexts. While the RLKWiC dataset represents a step towards satisfying Gon{\c{c}}alves'~\cite{gonccalves2011pseudo} vision, it is still specific to context-related tasks and affected by data incompleteness due to censoring required to preserve privacy and confidentiality. 

Given the challenges of collecting comprehensive, heavily annotated datasets and their persistent issues with adaptability, generalization, scalability, and data incompleteness due to censoring, we propose KnoWoGen, a knowledge work dataset generator. KnoWoGen simulates multiple knowledge workers solving tasks, creating documents, and collaborating. While a Multi-agent System handles the task distribution and scheduling, documents are generated by prompting Large Language Models. Moreover, our solution enables engineers of knowledge work assistance tools to configure and create a dataset fitting their evaluation needs. Additionally, resulting datasets can be publicly shared for comparison. All background information given in the configuration or introduced during the simulation is stored in a knowledge graph to preserve all contextually relevant information that can be later used as input, ground truth information, or for analysis purposes when evaluating assistance tools on the generated dataset. 

This paper is organized as follows: In Section \ref{related_work}, we discuss related work. In Section \ref{methodology}, we present the design of our KnoWoGen approach. Section \ref{experiments} examines the gap between generated and real knowledge work documents and provides an in-depth analysis of decisive criteria. Section \ref{discussion} discusses the experiment results and derived improvement potentials. Finally, Section \ref{conclusion} summarizes our findings and outlines potential future work.

\section{Related Work} 
\label{related_work}
There are multiple generators related to our approach. Guo et al.~\cite{guo2005lubm} introduced the knowledge graph generator LUBM, a customizable, scalable solution for triple store benchmarking. Other generators target specific document types related to knowledge work, such as spreadsheets~\cite{DBLP:conf/esws/SchroderJD21} or emails~\cite{DBLP:conf/bigdataconf/BabalolaJUD18}. The approach of Schulze et al.~\cite{DBLP:conf/semweb/SchulzeSJ021} is also similar to the KnoWoGen. It simulates business processes resulting in accounting documents, such as invoices, and a knowledge graph modeling the processes and associations with documents. This approach and the previously introduced document generators have in common that they synthesize documents by modifying examples or working with blueprints and template patterns. Consequently, the diversity of generated documents is limited. Recently, with the emergence of Large Language Models (LLMs), new document generators evolved based on them (see, e.g., ~\cite{DBLP:journals/corr/abs-2301-00665,DBLP:conf/eacl/MondalSNGBB24}) which, in contrast to earlier solutions, can generate more diverse documents without requiring explicit templates or examples.

Moreover, several simulators based on LLM-operated Multi-agent Systems emerged. Wang et al.~\cite{wang2024survey} and Guo et al.~\cite{guo2024large} provide an overview of these approaches. In contrast to typical Multi-agent Systems, the agents' subsequent actions are not decided randomly or by following a predefined concrete strategy with a set of rules. Instead, the LLM decides and executes the following actions.

Our approach also utilizes LLMs to generate documents reflecting the task-solving actions of one or multiple knowledge workers. However, the LLM is not responsible for deciding which successive actions an agent will perform.
Nevertheless, the generated documents influence the outcome of subsequent actions in which, for instance, a document is synthesized based on the content of a previous one. 
Additionally, our tool generates a corresponding knowledge graph that, amongst others, also puts the generated documents in relation. 

Since we wanted to evaluate how authentic generated knowledge work documents are perceived by humans and which differences to real documents exist, we conducted an experiment inspired by the Turing test \cite{DBLP:journals/x/Turing50}. Hence, we follow the examples of, for instance, Jannai et al. \cite{jannai2023human} who examined whether humans can distinguish human- and AI chat partners, Clark et al. \cite{clark-etal-2021-thats} conducting a similar test on recipes, stories, and news articles and Jakesch et al. \cite{jakesch2023human} analyzing generated and real self-introductions.

\section{Methodology}
\label{methodology}
This section introduces the design of our knowledge work dataset generator KnoWoGen. First, we provide an overview and then explain the simulation setup and process in more detail. 

\paragraph{Overview:}
KnoWoGen is a generator framework that synthesizes knowledge work datasets by simulating the collaborative knowledge work of multiple agents.
A knowledge work assistance tool engineer needs to configure the KnoWoGen accordingly to create a suitable dataset to evaluate or optimize their tool.
Based on these settings, KnoWoGen sets up the simulation environment with agents and tasks. Subsequently, in each simulation round, agents get tasks assigned and conduct them, resulting in either documents or data traces. The KnoWoGen stores all information about the simulation setting and completed tasks in a knowledge graph.

\paragraph{Simulating Knowledge Workers and Their Environment:}
The configuration sets up the simulation environment. It encompasses the specification of agents, companies, departments, and the domain. In the current implementation, agents are defined by their names and job roles. However, they can also be specified in terms of their behavior modeled as a set of rules, e.g., whether an agent usually responds to emails very briefly or circumlocutory, or by relationships to others, e.g., who is the boss of whom. These properties should be utilized for corresponding actions executed by the knowledge worker and for the task assignment.

Moreover, the users can specify general simulation settings in advance, e.g., how many tasks the agents should complete or the probability that an agent will get sick. In the current prototype, the focus was on defining tasks to shape the simulation. In the next subsection, we explain how tasks are designed and how they can be specified by the KnoWoGen users.  

\paragraph{Tasks:}
Tasks are a central element of the simulation and specify the activities of the knowledge workers. In our design, tasks are sequences of actions with logical or content dependencies, i.e., a meeting invitation has to be written and sent out before the actual meeting happens. Examples of tasks are, for instance, preparing for a workshop or writing a paper. Hence, tasks comprise multiple activities, here called actions, like writing a section for the paper or giving feedback to a colleague's paper. We base our definitions of action types on the taxonomy of Reinhardt et al. \cite{Reinhardt2011KnowledgeWR} defining types such as authoring, dissemination, and information search. 

Users of the KnoWoGen have to specify the tasks in the configuration \footnote{In the current prototype, the configuration can be specified in a TOML file}. Tasks are defined by their frequency, which actions they comprise, and other general properties relevant to all actions, like the topic of the task, which agents should be involved, and other involved entities. For instance, a project or product subject to a task can be defined. Actions are characterized by their duration, their action type, and other type-specific properties, e.g., for an authoring task, the document type that should be written must be set. Moreover, pointers to actions, a certain action depends on, have to be given such that these dependencies can be considered by the simulation engine. 

The tasks guide the simulation process. In contrast to other typical Multi-agent Systems (MAS), the actions are predefined and not chosen based on environmental observation. Nevertheless, the output of actions still shapes the environment, and previously generated documents can be considered when producing others. For specific actions, generated documents can be analyzed and follow-up actions can be derived. For instance, this is the case for email conversations since an email can contain questions that should be answered by the receiver and should spark an action of writing a response.
Another difference to other MAS is that there is also no assessment phase of the actions' outputs because the objective of the KnoWoGen is not necessarily optimally solving knowledge work tasks. Instead, the goal is to generate authentic documents and data traces which are difficult to check automatically.  

Most actions result in a document since the desired knowledge work dataset should be document-centric.
There are, however, also actions that only result in data traces, e.g., when putting a document in a folder or searching.

\paragraph{Generating Documents:}
Documents are created when knowledge workers execute actions assigned to them. They represent the outcome of completed actions. Depending on the action configuration, a prompt is composed and sent to an instruction-fine-tuned LLM that generates a suitable document based on the given instructions. 
Hereby, relevant parameters of the simulation environment are utilized as input, e.g., which agents are involved in the action, which topic should be discussed, and which document type should be generated. Those inputs either stem from the task configuration or are randomly sampled during the simulation setup.
The prompts are composed of up to four parts that are listed and described in the following \footnote{Concrete examples of prompts and corresponding, generated documents can be found here: \url{https://purl.archive.org/knowogen/examples}}:

\begin{itemize}
    \item A system prompt part describes the general goal of generating an authentic artificial HTML-formatted document that might be filled with additional information not mentioned in the subsequent prompt parts.
    \item Depending on the action, an instruction part specific to the underlying action type and configuration, states, for instance, which agents are involved, which topic the document should address, and which type of document should be generated. 
    \item Summaries of previous documents that should be considered are included.
    \item Further document type-specific instructions targeting problems found in pretests.
\end{itemize}

The choice of the concrete LLM connected to the KnoWoGen is part of the configuration. LLMs with higher context length limits are preferable since the instructions can be more detailed, and longer documents can be synthesized with one prompt. Depending on the chosen LLM, prompt templates tailored to a different LLM might have to be adapted minorly to achieve comparable results.
\paragraph{Filling the Knowledge Graph:}
Besides the environment setup, the contextual information of the simulation steps is stored in the knowledge graph. This includes, in particular, all details about tasks, e.g., all executed actions and their properties like their dependencies, involved agents, and other parameters. Moreover, all resulting data traces and documents are included and connected to their underlying actions.

The simulation context stored in the knowledge graph can later be used as input or ground truth depending on the tools that should be optimized or evaluated. For example, task information can be used as ground truth for a task predictor. Similarly, document content dependencies can be used for corresponding predictors or action parameters for classifiers.

\section{Experiments}
\label{experiments}
Similar to previous publications \cite{clark-etal-2021-thats,jakesch2023human,jannai2023human}, we examined how authentic generated documents, in this case knowledge work documents, are perceived by human evaluators compared to real ones. With this experiment, we assessed our approach's document generation capability. Document generation was, in the past, one main challenge of dataset generators and is an important factor for the general feasibility of our approach. Moreover, participants were asked in the experiment to explain their ratings to get more insights into the driving factors of their judgment and essential aspects of document authenticity.

\paragraph{Setup:}
The experiment dataset was composed of 25 emails and meeting minutes. We chose these two document types since they reflect or represent collaborative knowledge work. Hence, compared to other documents such as papers or project reports, they are associated with higher authenticity requirements since the collaboration or communication captured in the documents must also appear human-like.

To increase comparability, we selected five content categories for the included documents. Emails were either meeting invitations, retrospective meeting-related exchanges, or feedback on an external document. Included meeting minutes were discussions about the current work status. One category of meeting minutes involved also planning, i.e., a distribution of future tasks, and the other one focused solely on exchanging the status.

For each category, we collected one real example and generated four examples with the Llama-13B-Chat model~\cite{touvron2023llama}. The real emails were randomly selected from the Enron dataset~\cite{DBLP:conf/ecml/KlimtY04} and the meeting minutes from the ELITR dataset~\cite{DBLP:conf/lrec/NedoluzhkoSHGB22}.
Since we wanted to test two generation variables and for each one two different variants, we decided on a multivariate experiment.
Generated documents were either more greedy or more deterministic and generated in a zero-shot or a two-shot manner, i.e., providing no or two examples of real documents in the prompt. For the two-shot variants, we also randomly sampled further examples from the two aforementioned datasets.
All synthesized documents resulted from small simulations, i.e., they were the result of either the first or the second simulation step and accordingly only relied on at most one previously generated document.

We asked participants to rate on a 7-point Likert scale~\cite{likert1932technique} how realistic the documents appeared to them. Additionally, they were encouraged to justify their decision. However, this was optional to avoid forcing participants to explain their decision when they were unsure about concrete influence factors. For each participant, documents were randomly ordered to decrease the effects of their order. Moreover, they were not told about the distribution of real and generated documents to not bias their judgments.

In total, 29 participants, aged between 18 and 64, rated the given knowledge work documents. Among them, most had a computer science background. Predominately, they were researchers (38\%), students (31\%), and software engineers (17\%). 87\% assessed their English proficiency as being on a B2 or C1 level
~\footnote{\raggedright According to the Common European Framework of Reference for Languages (CEFR) (https://www.europaeischer-referenzrahmen.de/)}. 
Regarding their acquaintance with LLMs, the participants were divided almost equally into regular and irregular LLM users.

\paragraph{Results:}
53\% of the generated documents received an authenticity rating between five and seven on the Likert scale meaning that they are assessed as rather authentic to very authentic. For comparison, 74\% of the real documents got a score in the same range assigned.\footnote{Complete score distribution for real documents: (1: 4\%, 2: 5\%, 3: 6\%, 4: 11\%, 5: 16\%, 6: 28\%, 7: 31\%) and for generated ones: (1: 8\%, 2: 14\%, 3: 16\%, 4: 9\%, 5: 15\%, 6: 17\%, 7: 20\%). The KL divergence of the two distributions is 15.91\%. A visualization can be found at the KnoWoGen website: \url{https://purl.archive.org/knowogen/document_authenticity_experiment}}. 
There was no noticeable difference in the ratings between regular and irregular users or between the different generation configurations (i.e. between zero- vs. two-shot prompting and greedy vs. more deterministic sampling methods).

In total, participants commented on 451 of 725 document ratings. We have categorized the comments into different positive and negative aspects. A comprehensive table showing these categorizations can be found in the appendix. The most frequent negative points of criticism were the lack of details and too generic contents, unauthentic values for addresses, phone numbers, and other named entities, and repetitiveness regarding words or formulations. Positive comments were predominately about the inclusion of details, a good overall structure, and an authentic writing style with imperfections like spelling or grammar mistakes as well as incomplete sentences or parentheses.

\section{Discussion}
\label{discussion}
Overall, the results indicate that the KnoWoGen can generate authentic documents. Still, if the detected issues are resolved, an even higher authenticity is achievable. Moreover, the comments provided insights into which factors humans adduce to identify authentic documents which is valuable to optimize the generation.

Stylistic issues of generated documents, like small formatting issues or repetitive language, have a low impact on the objectives of the KnoWoGen since the produced datasets are intended for evaluating knowledge work support tools and not for Natural Language Processing tasks conducting a deep language analysis. In contrast, content-related issues are more important since content analysis is typical when aiming at supporting knowledge workers. For instance, for information extraction, search or recommendation. 

Regarding content-related issues, the lack of details or too generic content were among the most frequently mentioned problems.
Documents generated for this experiment were built based on a maximum of one previously generated document. A follow-up experiment could examine whether the situation improves when using documents generated based on more previously generated documents.
Besides, the prompts can be enhanced to encourage a more specific focus and more details which can likewise be achieved by a preceding step in which the focus aspects are generated. In some cases, aspects to focus on can also be predetermined. However, this would decrease the documents' diversity. 

Another typical problem was the generation of unauthentic names, addresses, or phone numbers that were filled in since this information was not part of the prompt. This could be solved by extracting such kinds of information for different document types from several generated documents and providing more authentic values explicitly in future prompts. 

Furthermore, the documents showed that the role of the author was not clear during the generation. In a few documents, the intended author given in the prompt did refer to themselves in the third person. Thus, in the future, the authorship must be made clearer including instructions stating that the author should refer to themselves in the first person.

Multiple other issues can be potentially solved by including more general instructions in the prompt, e.g., how to refer to colleagues or externals, keeping the document concise, or adding variations. Additionally, instructions could encourage characteristics that the experiment revealed as being particularly authentic, e.g., a few spelling or grammatical errors could make the resulting documents more realistic.

In summary, the experiment showed that the KnoWoGen can generate documents perceived as authentic. Moreover, the results indicate which aspects were most influential for the participants' judgments.
This information can be very beneficial for optimizing the authenticity of documents even more.
For future experiments, examining other document types, multiple related documents, and inter-document diversity would be interesting. In contrast, examining the knowledge graph or the simulation setting is not as meaningful as analyzing the generated documents as they reflect, by design, the customizable configuration.

\section{Conclusion}
\label{conclusion}
In this paper, we proposed KnoWoGen, a knowledge work dataset generator that can be tailored to different evaluation needs.
It targets common problems of existing data collections, like incompleteness, lack of background information, and limited applicability. The generator simulates multiple knowledge workers collaborating and completing tasks. As an outcome of most actions, documents are generated by prompting an LLM with the necessary contextual information. Our experiment showed that this approach is promising and indicated some optimization potential. In future works, document and task interdependencies could be increased, diversity in larger document collections could be analyzed and optimized, and synthesized documents could be used to influence the simulation more, e.g., by creating additional tasks derived from previously generated documents.

\section*{Acknowledgements} This work was funded by the German Federal Ministry of Education and Research (BMBF) in the project SensAI (grant no. 01IW20007).

\printbibliography
\clearpage
\appendix
\section*{Appendix}
\begin{table}[]
\resizebox{0.9\linewidth}{!}{%
\begin{tabular}{l|ccccccccc|}
\cline{2-10}
\textbf{} &
  \multicolumn{3}{c|}{\cellcolor[HTML]{C0C0C0}Generated} &
  \multicolumn{3}{c|}{\cellcolor[HTML]{C0C0C0}Real} &
  \multicolumn{3}{c|}{\cellcolor[HTML]{C0C0C0}Total} \\ \cline{2-10} 
 &
  \multicolumn{1}{l|}{\cellcolor[HTML]{EFEFEF}Low score} &
  \multicolumn{1}{l|}{\cellcolor[HTML]{EFEFEF}Neutral} &
  \multicolumn{1}{l|}{\cellcolor[HTML]{EFEFEF}High score} &
  \multicolumn{1}{l|}{\cellcolor[HTML]{EFEFEF}Low score} &
  \multicolumn{1}{l|}{\cellcolor[HTML]{EFEFEF}Neutral} &
  \multicolumn{1}{l|}{\cellcolor[HTML]{EFEFEF}High score} &
  \multicolumn{1}{l|}{\cellcolor[HTML]{EFEFEF}Low score} &
  \multicolumn{1}{l|}{\cellcolor[HTML]{EFEFEF}Neutral} &
  \multicolumn{1}{l|}{\cellcolor[HTML]{EFEFEF}High score} \\ \cline{2-10} 
 &
  \multicolumn{9}{c|}{\cellcolor[HTML]{FFCE93}\textbf{Number of Comments}} \\ \cline{2-10} 
 &
  \multicolumn{1}{c|}{210} &
  \multicolumn{1}{c|}{28} &
  \multicolumn{1}{c|}{149} &
  \multicolumn{1}{c|}{19} &
  \multicolumn{1}{c|}{9} &
  \multicolumn{1}{c|}{69} &
  \multicolumn{1}{c|}{229} &
  \multicolumn{1}{c|}{37} &
  218 \\ \cline{2-10} 
 &
  \multicolumn{9}{c|}{\cellcolor[HTML]{FFCE93}\textbf{Percentage of Comments}} \\ \hline
\multicolumn{1}{|c|}{\cellcolor[HTML]{FFCE93}\textbf{Categories}} &
  \multicolumn{9}{l|}{} \\ \hline
\multicolumn{1}{|l|}{\cellcolor[HTML]{FFCCC9}Plausibility (-)} &
  \multicolumn{1}{c|}{10\%} &
  \multicolumn{1}{c|}{11\%} &
  \multicolumn{1}{c|}{4\%} &
  \multicolumn{1}{c|}{16\%} &
  \multicolumn{1}{c|}{11\%} &
  \multicolumn{1}{c|}{1\%} &
  \multicolumn{1}{c|}{10\%} &
  \multicolumn{1}{c|}{11\%} &
  3\% \\ \hline
\multicolumn{1}{|l|}{\cellcolor[HTML]{FFCCC9}Correctness (-)} &
  \multicolumn{1}{c|}{1\%} &
  \multicolumn{1}{c|}{} &
  \multicolumn{1}{c|}{1\%} &
  \multicolumn{1}{c|}{} &
  \multicolumn{1}{c|}{} &
  \multicolumn{1}{c|}{} &
  \multicolumn{1}{c|}{1\%} &
  \multicolumn{1}{c|}{} &
  1\% \\ \hline
\multicolumn{1}{|l|}{\cellcolor[HTML]{FFCCC9}\begin{tabular}[c]{@{}l@{}}Format / Spelling /\\ Grammar Mistakes (-)\end{tabular}} &
  \multicolumn{1}{c|}{5\%} &
  \multicolumn{1}{c|}{} &
  \multicolumn{1}{c|}{6\%} &
  \multicolumn{1}{c|}{16\%} &
  \multicolumn{1}{c|}{} &
  \multicolumn{1}{c|}{4\%} &
  \multicolumn{1}{c|}{6\%} &
  \multicolumn{1}{c|}{} &
  6\% \\ \hline
\multicolumn{1}{|l|}{\cellcolor[HTML]{FFCCC9}Repetitive Wording (-)} &
  \multicolumn{1}{c|}{\textbf{16\%}} &
  \multicolumn{1}{c|}{4\%} &
  \multicolumn{1}{c|}{11\%} &
  \multicolumn{1}{c|}{11\%} &
  \multicolumn{1}{c|}{11\%} &
  \multicolumn{1}{c|}{4\%} &
  \multicolumn{1}{c|}{\textbf{16\%}} &
  \multicolumn{1}{c|}{5\%} &
  9\% \\ \hline
\multicolumn{1}{|l|}{\cellcolor[HTML]{FFCCC9}\begin{tabular}[c]{@{}l@{}}Exaggerated Enthusiam/\\ Optimistimism etc. (-)\end{tabular}} &
  \multicolumn{1}{c|}{5\%} &
  \multicolumn{1}{c|}{11\%} &
  \multicolumn{1}{c|}{5\%} &
  \multicolumn{1}{c|}{} &
  \multicolumn{1}{c|}{} &
  \multicolumn{1}{c|}{} &
  \multicolumn{1}{c|}{4\%} &
  \multicolumn{1}{c|}{8\%} &
  4\% \\ \hline
\multicolumn{1}{|l|}{\cellcolor[HTML]{FFCCC9}\begin{tabular}[c]{@{}l@{}}Unauthentic Social Aspects,\\ e.g., addressing people, \\ greeting (-)\end{tabular}} &
  \multicolumn{1}{c|}{9\%} &
  \multicolumn{1}{c|}{} &
  \multicolumn{1}{c|}{8\%} &
  \multicolumn{1}{c|}{5\%} &
  \multicolumn{1}{c|}{11\%} &
  \multicolumn{1}{c|}{} &
  \multicolumn{1}{c|}{9\%} &
  \multicolumn{1}{c|}{3\%} &
  6\% \\ \hline
\multicolumn{1}{|l|}{\cellcolor[HTML]{FFCCC9}Unnecessary Information (-)} &
  \multicolumn{1}{c|}{7\%} &
  \multicolumn{1}{c|}{4\%} &
  \multicolumn{1}{c|}{5\%} &
  \multicolumn{1}{c|}{} &
  \multicolumn{1}{c|}{} &
  \multicolumn{1}{c|}{} &
  \multicolumn{1}{c|}{7\%} &
  \multicolumn{1}{c|}{3\%} &
  3\% \\ \hline
\multicolumn{1}{|l|}{\cellcolor[HTML]{FFCCC9}\begin{tabular}[c]{@{}l@{}}Generic Content / Lack of \\ Details (-)\end{tabular}} &
  \multicolumn{1}{c|}{\textbf{32\%}} &
  \multicolumn{1}{c|}{\textbf{21\%}} &
  \multicolumn{1}{c|}{6\%} &
  \multicolumn{1}{c|}{5\%} &
  \multicolumn{1}{c|}{} &
  \multicolumn{1}{c|}{4\%} &
  \multicolumn{1}{c|}{\textbf{30\%}} &
  \multicolumn{1}{c|}{\textbf{16\%}} &
  6\% \\ \hline
\multicolumn{1}{|l|}{\cellcolor[HTML]{FFCCC9}\begin{tabular}[c]{@{}l@{}}Unauthentic Names /Numbers, \\ e.g.,Address, phone numbers etc.\\ (-)\end{tabular}} &
  \multicolumn{1}{c|}{\textbf{17\%}} &
  \multicolumn{1}{c|}{14\%} &
  \multicolumn{1}{c|}{7\%} &
  \multicolumn{1}{c|}{} &
  \multicolumn{1}{c|}{} &
  \multicolumn{1}{c|}{1\%} &
  \multicolumn{1}{c|}{\textbf{15\%}} &
  \multicolumn{1}{c|}{11\%} &
  5\% \\ \hline
\multicolumn{1}{|l|}{\cellcolor[HTML]{FFCCC9}Author Identity Problems (-)} &
  \multicolumn{1}{c|}{6\%} &
  \multicolumn{1}{c|}{4\%} &
  \multicolumn{1}{c|}{3\%} &
  \multicolumn{1}{c|}{} &
  \multicolumn{1}{c|}{} &
  \multicolumn{1}{c|}{} &
  \multicolumn{1}{c|}{6\%} &
  \multicolumn{1}{c|}{3\%} &
  2\% \\ \hline
\multicolumn{1}{|l|}{\cellcolor[HTML]{FFCCC9}Repetitive Content (-)} &
  \multicolumn{1}{c|}{3\%} &
  \multicolumn{1}{c|}{7\%} &
  \multicolumn{1}{c|}{2\%} &
  \multicolumn{1}{c|}{} &
  \multicolumn{1}{c|}{} &
  \multicolumn{1}{c|}{} &
  \multicolumn{1}{c|}{3\%} &
  \multicolumn{1}{c|}{5\%} &
  1\% \\ \hline
\multicolumn{1}{|l|}{\cellcolor[HTML]{FFCCC9}Badly Written / Structured (-)} &
  \multicolumn{1}{c|}{4\%} &
  \multicolumn{1}{c|}{11\%} &
  \multicolumn{1}{c|}{2\%} &
  \multicolumn{1}{c|}{\textbf{53\%}} &
  \multicolumn{1}{c|}{} &
  \multicolumn{1}{c|}{3\%} &
  \multicolumn{1}{c|}{8\%} &
  \multicolumn{1}{c|}{8\%} &
  2\% \\ \hline
\multicolumn{1}{|l|}{\cellcolor[HTML]{FFCCC9}Unauthentic Language / Style (-)} &
  \multicolumn{1}{c|}{3\%} &
  \multicolumn{1}{c|}{4\%} &
  \multicolumn{1}{c|}{} &
  \multicolumn{1}{c|}{5\%} &
  \multicolumn{1}{c|}{} &
  \multicolumn{1}{c|}{} &
  \multicolumn{1}{c|}{3\%} &
  \multicolumn{1}{c|}{3\%} &
   \\ \hline
\multicolumn{1}{|l|}{\cellcolor[HTML]{FFCCC9}Use of Emojis (-)} &
  \multicolumn{1}{c|}{4\%} &
  \multicolumn{1}{c|}{4\%} &
  \multicolumn{1}{c|}{3\%} &
  \multicolumn{1}{c|}{} &
  \multicolumn{1}{c|}{} &
  \multicolumn{1}{c|}{} &
  \multicolumn{1}{c|}{4\%} &
  \multicolumn{1}{c|}{3\%} &
  2\% \\ \hline
\multicolumn{1}{|l|}{\cellcolor[HTML]{FFCCC9}\begin{tabular}[c]{@{}l@{}}Too Consistent,\\ e.g., one phrase for every aspect\\ (-)\end{tabular}} &
  \multicolumn{1}{c|}{3\%} &
  \multicolumn{1}{c|}{4\%} &
  \multicolumn{1}{c|}{1\%} &
  \multicolumn{1}{c|}{} &
  \multicolumn{1}{c|}{} &
  \multicolumn{1}{c|}{1\%} &
  \multicolumn{1}{c|}{3\%} &
  \multicolumn{1}{c|}{3\%} &
  1\% \\ \hline
\multicolumn{1}{|l|}{\cellcolor[HTML]{FFCCC9}Missing Parts / Information (-)} &
  \multicolumn{1}{c|}{7\%} &
  \multicolumn{1}{c|}{7\%} &
  \multicolumn{1}{c|}{3\%} &
  \multicolumn{1}{c|}{} &
  \multicolumn{1}{c|}{\textbf{22\%}} &
  \multicolumn{1}{c|}{3\%} &
  \multicolumn{1}{c|}{6\%} &
  \multicolumn{1}{c|}{11\%} &
  3\% \\ \hline
\multicolumn{1}{|l|}{\cellcolor[HTML]{FFCCC9}Inconsistencies (-)} &
  \multicolumn{1}{c|}{1\%} &
  \multicolumn{1}{c|}{4\%} &
  \multicolumn{1}{c|}{1\%} &
  \multicolumn{1}{c|}{5\%} &
  \multicolumn{1}{c|}{\textbf{22\%}} &
  \multicolumn{1}{c|}{3\%} &
  \multicolumn{1}{c|}{2\%} &
  \multicolumn{1}{c|}{8\%} &
  2\% \\ \hline
\multicolumn{1}{|l|}{\cellcolor[HTML]{FFCCC9}"Typical LLM Phrases" (-)} &
  \multicolumn{1}{c|}{4\%} &
  \multicolumn{1}{c|}{} &
  \multicolumn{1}{c|}{} &
  \multicolumn{1}{c|}{} &
  \multicolumn{1}{c|}{} &
  \multicolumn{1}{c|}{} &
  \multicolumn{1}{c|}{4\%} &
  \multicolumn{1}{c|}{} &
   \\ \hline
\multicolumn{1}{|l|}{\cellcolor[HTML]{BBF0BA}\begin{tabular}[c]{@{}l@{}}Authentic Social Aspects,\\ e.g., addressing people, \\ greeting (+)\end{tabular}} &
  \multicolumn{1}{c|}{1\%} &
  \multicolumn{1}{c|}{} &
  \multicolumn{1}{c|}{5\%} &
  \multicolumn{1}{c|}{} &
  \multicolumn{1}{c|}{} &
  \multicolumn{1}{c|}{13\%} &
  \multicolumn{1}{c|}{1\%} &
  \multicolumn{1}{c|}{} &
  8\% \\ \hline
\multicolumn{1}{|l|}{\cellcolor[HTML]{BBF0BA}Well Written (Concise, clear) (+)} &
  \multicolumn{1}{c|}{1\%} &
  \multicolumn{1}{c|}{} &
  \multicolumn{1}{c|}{3\%} &
  \multicolumn{1}{c|}{} &
  \multicolumn{1}{c|}{} &
  \multicolumn{1}{c|}{1\%} &
  \multicolumn{1}{c|}{1\%} &
  \multicolumn{1}{c|}{} &
  3\% \\ \hline
\multicolumn{1}{|l|}{\cellcolor[HTML]{BBF0BA}Authentic Language / Style (+)} &
  \multicolumn{1}{c|}{2\%} &
  \multicolumn{1}{c|}{4\%} &
  \multicolumn{1}{c|}{\textbf{18\%}} &
  \multicolumn{1}{c|}{} &
  \multicolumn{1}{c|}{} &
  \multicolumn{1}{c|}{\textbf{38\%}} &
  \multicolumn{1}{c|}{2\%} &
  \multicolumn{1}{c|}{3\%} &
  \textbf{24\%} \\ \hline
\multicolumn{1}{|l|}{\cellcolor[HTML]{BBF0BA}Well Structured (+)} &
  \multicolumn{1}{c|}{1\%} &
  \multicolumn{1}{c|}{4\%} &
  \multicolumn{1}{c|}{6\%} &
  \multicolumn{1}{c|}{} &
  \multicolumn{1}{c|}{} &
  \multicolumn{1}{c|}{} &
  \multicolumn{1}{c|}{1\%} &
  \multicolumn{1}{c|}{3\%} &
  4\% \\ \hline
\multicolumn{1}{|l|}{\cellcolor[HTML]{BBF0BA}Details Included} &
  \multicolumn{1}{c|}{} &
  \multicolumn{1}{c|}{4\%} &
  \multicolumn{1}{c|}{10\%} &
  \multicolumn{1}{c|}{} &
  \multicolumn{1}{c|}{} &
  \multicolumn{1}{c|}{\textbf{16\%}} &
  \multicolumn{1}{c|}{} &
  \multicolumn{1}{c|}{3\%} &
  12\% \\ \hline
\multicolumn{1}{|l|}{\cellcolor[HTML]{BBF0BA}Emotions Reflected (+)} &
  \multicolumn{1}{c|}{} &
  \multicolumn{1}{c|}{} &
  \multicolumn{1}{c|}{5\%} &
  \multicolumn{1}{c|}{} &
  \multicolumn{1}{c|}{} &
  \multicolumn{1}{c|}{} &
  \multicolumn{1}{c|}{} &
  \multicolumn{1}{c|}{} &
  4\% \\ \hline
\multicolumn{1}{|l|}{\cellcolor[HTML]{BBF0BA}Spelling / Grammar Mistakes (+)} &
  \multicolumn{1}{c|}{1\%} &
  \multicolumn{1}{c|}{4\%} &
  \multicolumn{1}{c|}{4\%} &
  \multicolumn{1}{c|}{5\%} &
  \multicolumn{1}{c|}{} &
  \multicolumn{1}{c|}{\textbf{16\%}} &
  \multicolumn{1}{c|}{2\%} &
  \multicolumn{1}{c|}{3\%} &
  8\% \\ \hline
\multicolumn{1}{|l|}{\cellcolor[HTML]{BBF0BA}Formatting Mistakes (+)} &
  \multicolumn{1}{c|}{} &
  \multicolumn{1}{c|}{} &
  \multicolumn{1}{c|}{} &
  \multicolumn{1}{c|}{} &
  \multicolumn{1}{c|}{} &
  \multicolumn{1}{c|}{12\%} &
  \multicolumn{1}{c|}{} &
  \multicolumn{1}{c|}{} &
  4\% \\ \hline
\multicolumn{1}{|l|}{\cellcolor[HTML]{BBF0BA}Realistic Names (+)} &
  \multicolumn{1}{c|}{} &
  \multicolumn{1}{c|}{4\%} &
  \multicolumn{1}{c|}{2\%} &
  \multicolumn{1}{c|}{} &
  \multicolumn{1}{c|}{} &
  \multicolumn{1}{c|}{3\%} &
  \multicolumn{1}{c|}{} &
  \multicolumn{1}{c|}{3\%} &
  2\% \\ \hline
\end{tabular}%
}
\caption{
This table shows the percentage of comments for different settings (real/generated and low/neutral/high score) that fit into the given categories. Since some comments were ambiguous or mentioned aspects that in total less than four others also commented, not all comments were assigned to the shown categories. Additionally, several comments are assigned to multiple categories. The color and +/- symbols of the different categories show whether the aspect influenced the rating positively (green, +) or negatively (red, -). Empty cells indicate that no comment of the setting was assigned to the category. 
The table can be read as follows: The first cell on the left-hand side with the value 10\% indicates that 10\% of the comments accompanying low ratings of generated documents state that plausibility problems occurred in the documents.
}
\end{table}
\end{document}